\documentclass[conference]{IEEEtran}
\IEEEoverridecommandlockouts
\usepackage{amsmath,amssymb,amsfonts}
\usepackage{cite}
\usepackage{multirow}
\usepackage[utf8]{inputenc}
\usepackage{amssymb}
\usepackage{marvosym}
\usepackage{wasysym}
\usepackage{algorithmic}
\usepackage{graphicx}
\usepackage{textcomp}
\usepackage{xcolor}
\usepackage[normalem]{ulem}
\usepackage{caption}
\useunder{\uline}{\ul}{}
\def\BibTeX{{\rm B\kern-.05em{\sc i\kern-.025em b}\kern-.08em
    T\kern-.1667em\lower.7ex\hbox{E}\kern-.125emX}}
\begin{document}

\title{DA-HFNet: Progressive Fine-Grained Forgery Image Detection and Localization Based on Dual Attention\\
}
\makeatletter
\newcommand{\linebreakand}{%
\end{@IEEEauthorhalign}
\hfill\mbox{}\par
\mbox{}\hfill\begin{@IEEEauthorhalign}
}
\makeatother

\author{\IEEEauthorblockN{1\textsuperscript{st} Yang Liu}
\IEEEauthorblockA{\textit{Laboratory for Big Data and Decision}\\
\textit{National University of Defense Technology}\\
ChangSha, China \\
liuyang.nudt@nudt.edu.cn}
\\
\IEEEauthorblockN{2\textsuperscript{nd} Xiaofei Li}
\IEEEauthorblockA{\textit{Laboratory for Big Data and Decision} \\
\textit{National University of Defense Technology}\\
Changsha, China \\
 xf@nudt.edu.cn}
\\
\IEEEauthorblockN{3\textsuperscript{rd} Jun Zhang}
\IEEEauthorblockA{\textit{Laboratory for Big Data and Decision} \\
\textit{National University of Defense Technology}\\
Changsha, China \\
zhangjun1975@nudt.edu.cn}
~\\
\and
\IEEEauthorblockN{4\textsuperscript{th} Shengze Hu}
\IEEEauthorblockA{\textit{Laboratory for Big Data and Decision} \\
\textit{National University of Defense Technology}\\
Changsha, China \\
springsun@nudt.edu.cn}
~\\
\IEEEauthorblockN{5\textsuperscript{th} Jun Lei*}
\IEEEauthorblockA{\textit{Laboratory for Big Data and Decision} \\
\textit{National University of Defense Technology}\\
Changsha, China \\
534205986@qq.com}
*Corresponding author
~\\
}

\maketitle

\begin{abstract}
The increasing difficulty in accurately detecting forged images generated by AIGC(Artificial Intelligence Generative Content) poses many risks, necessitating the development of effective methods to identify and further locate forged areas. In this paper, to facilitate research efforts, we construct a DA-HFNet forged image dataset guided by text or image-assisted GAN and Diffusion model. Our goal is to utilize a hierarchical progressive network to capture forged artifacts at different scales for detection and localization. Specifically, it relies on a dual-attention mechanism to adaptively fuse multi-modal image features in depth, followed by a multi-branch interaction network to thoroughly interact image features at different scales and improve detector performance by leveraging dependencies between layers. Additionally, we extract more sensitive noise fingerprints to obtain more prominent forged artifact features in the forged areas. Extensive experiments validate the effectiveness of our approach, demonstrating significant performance improvements compared to state-of-the-art methods for forged image detection and localization.The code and dataset will be released in the future.
\end{abstract}


\section{Introduction}
Image forgery detection and localization have always been key research topics in the field of artificial intelligence security. For a forged image, people not only want to detect that it is forged but also want to further refine the localization of which areas in the image are forged. With the remarkable performance of AI drawing and large-scale language models like ChatGPT, the concept of has become well-known, enabling the generation of realistic images. Currently, methods in the image generation field can be roughly categorized into three types of generative models: Variational Autoencoder (VAE)\cite{name1}, Generative Adversarial Network (GAN)\cite{name2}, and Diffusion Model (DM)\cite{name3}. They can learn the distribution of images from a large number of training images and generate similar images. It is worth noting that earlier generative models learned the distribution space of a large number of images to generate a complete image, where all pixels in these images are forged pixels, such as DDPM\cite{name4}. Nowadays, generative models have evolved to be able to perform partial edits on images, known as "Inpainting," for example, Inpaint Anything\cite{name5}. This method allows real pixels and forged pixels to coexist in the same image, posing more challenges to distinguishing between real and forged images.

Although generative models have brought many conveniences to areas such as image editing\cite{name6}, image restoration\cite{name7}, and image fusion\cite{name8}, when AI-altered images are maliciously exploited, humans cannot differentiate between real and forged images with the naked eye, and traditional detection models struggle to make accurate judgments, potentially causing immeasurable losses of false information to individuals and society. Unlike traditional image tampering methods such as copying, pasting, and moving, in recent years, large-scale image generation models have been applied to image editing, resulting in higher-quality generated images and increasingly blurred boundaries of tampered areas, significantly increasing the difficulty of forgery detection and localization. In addition to legal constraints, it is more important to improve the detection and localization capabilities of forged images at the technical level.

To address the challenges posed by high-quality forged images and new forgery methods to the identification work, we have studied a progressive hierarchical network for more refined forgery image detection and localization. First, due to the lack of image editing-related generated content in the current AIGC-generated image datasets, we constructed a DA-HFNet dataset covering whole-image generation, partial editing, and real images for research purposes. It includes GAN and DM generation methods guided by text or images. To our knowledge, this is the first AI-generated image dataset that includes both text-guided and image-guided image editing. We provide some examples of forged images in our dataset, as shown in Figure 1.

\begin{figure}[h]
	\centering
	\includegraphics[width=\linewidth]{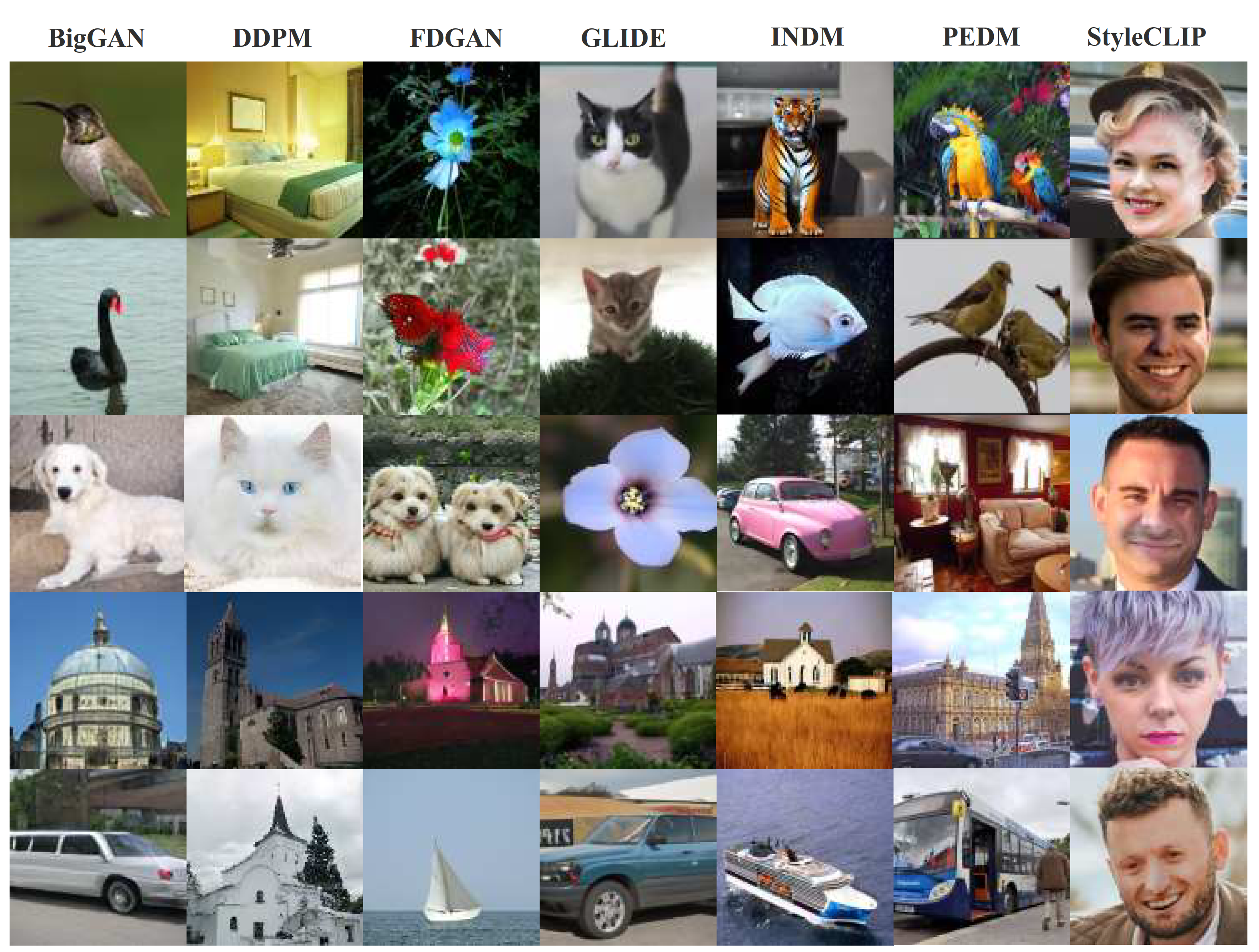}
	\caption{DA-HFNet Synthetic Image Dataset Examples.It is a dataset that is based on both GAN and diffusion models, encompassing both completely fabricated and partially edited images.}
\end{figure}

Although AIGC-generated images can achieve realistic effects, traces of the generation process still remain within them. This is because both GAN and DM rely on learning the distribution of images and generating images based on their own parameters and weights. For example, in previous studies\cite{name9,name10,name11,name12,name13}, researchers found various fingerprint features left inside synthetic images generated by generative models. For instance, Huang et al\cite{name12}. used manipulation traces extracted from RGB flows and local noise extracted from Noise flows for precise localization of image tampering. At the same time, other researchers\cite{name11,name14} found significant differences in frequency domain features between different types of forged images. Similar differences can positively impact the detection capabilities of models. Intuitively, to improve the model's detection capabilities for different models, it is necessary to explore stronger fingerprint features that can reflect GANs and diffusion models and their variants. Therefore, we can train an extractor to explore forged traces in images to accurately detect and localize forged images.

Currently, many researchers\cite{name10,name12,name15} simultaneously use various image features to achieve the discrimination of forged images and have achieved good results. Therefore, we also utilize multiple image features and strengthen the fusion of features. To learn richer feature representations, we perform dual-attention adaptive fusion on features of different categories. Specifically, we fuse RGB features in the image spatial domain and noise features, while using frequency domain features as a supervised branch. We then weight the spatial domain features and frequency domain features to enhance the expressive power of features.

Finally, since forged regions in images often have significant differences in position and scale, a fixed-resolution feature map may overlook some forged features in certain regions. The scale variation of forged regions will negatively affect the performance of the detector, and most existing works have ignored this point. Therefore, we fully interact features of different resolutions and use a hierarchical network to perform detection and localization of images at different scales.In addition, we also use a kind of edge loss to correct the location of the forgery area, which further improves the performance of the model.

Contributions of this paper:
\begin{itemize}
	\item Constructed a DA-HFNet forged image dataset covering current mainstream generative models, including text-guided image editing and image-guided image editing.
	\item Utilized a dual-attention mechanism for deeper feature fusion and multi-scale feature interaction in a multi-branch network.
	\item The experiment verifies that the proposed progressive network can effectively detect AIGC-forged images, especially the recent high-quality generated images accurately.
\end{itemize}

\section{Related Work}

Forgery Image Generation.Image generation models aim to learn the distribution of training samples from given image data and sample similar images. The current mainstream image generation models are mostly based on VAE, GAN, and Diffusion Model. Different generation methods can be roughly categorized into two types. One type is the whole-image generation mode, where these methods can generate complete forged images using images as input, such as CycleGAN\cite{name16}, LinkGAN\cite{name17}, DDPM\cite{name4}, as well as generating a complete forged image using text as input, such as StyleGAN\cite{name18}, GigaGAN\cite{name19}, GLIDE\cite{name20}, Stable Diffusion\cite{name21}, etc. The other type is the partial editing mode, where these methods can partially tamper with the target image guided by images or text, such as EditGAN\cite{name22}, DragGAN\cite{name23}, HD-painter\cite{name7}, Paint-by-example\cite{name24}, and manipulating a part of the input image guided by text, such as StyleCLIP\cite{name25}, Imagic\cite{name6}, Ranni\cite{name26}, etc. Image tampering based on generative models differs from traditional splicing methods because real and forged pixels coexist in the same image, and the forged boundaries are more concealed. Moreover, methods guided by text lack mask signals for supervision, making it more challenging for models to successfully detect and locate such images.

Forgery Image Detection.The goal of forgery image identification is not only to distinguish between real and forged images but also to further locate the forged areas within forged images. Previous research\cite{name9,name10,name27,name28,name29} aimed to extract tampering traces from forged images for binary classification tasks and achieved good results in traditional image tampering methods. However, when the training and testing sets are extended to include forged images generated by generative models, the performance of traditional methods significantly decreases. To detect AIGC-generated images, researchers have conducted extensive explorations. Sha et al.\cite{name30} captured the characteristic that the correlation between images and text in generated images is greater than that in real images and used the correlation between image-text pairs as detection features. Xi et al.\cite{name13} combined multiple spatial domain features and designed a dual-stream detection network. Wu et al.\cite{name31} targeted the characteristic that the information contained in text-guided generated images does not exceed the text prompt and developed a detection method for generated images. Researchers have also explored new fingerprints. Wang et al.\cite{name32} found that the error of diffusion model-generated images after diffusion reconstruction is smaller than that of real images. Zhong et al. \cite{name14}found that pixels in complex texture regions of AIGC-generated images exhibit more significant fluctuations than those in flat texture regions. Guillaro et al.\cite{name10} learned a noise-sensitive fingerprint to extract high-level and low-level fingerprints of generated images. David et al. \cite{name33}attempted to address emerging new generation methods by simulating the order disclosed by generative models for online learning.In summary, the current forgery detection methods suffer from limitations in forgery feature extraction, resulting in inadequate detection and localization of forged images.

\begin{figure*}[h]
	\centering
	\includegraphics[width=\linewidth]{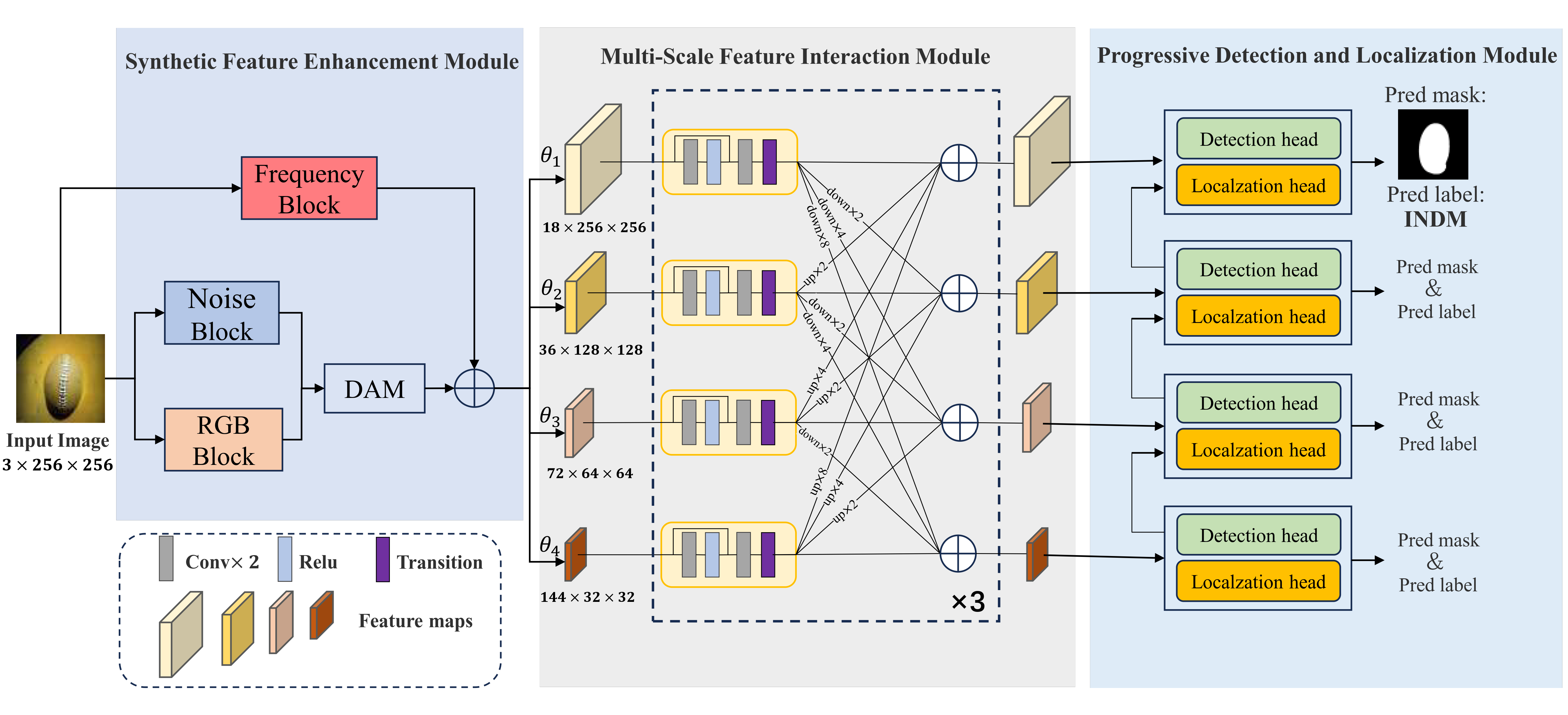}
	\caption{DA-HFNet Model Structure.}
\end{figure*}

\section{Proposed Method}
Our goal is to extract hidden artifacts in forged images and utilize a hierarchical network to improve the model's detection and localization capabilities for forged images. In this section, we will introduce DA-HFNet, as shown in Figure 2, which consists of three modules: Forgery Feature Enhancement Module (Section 3.1), Multi-Scale Feature Interaction Module (Section 3.2), and Progressive Detection and Localization Module (Section 3.3).Given an input image, the feature extraction and fusion take place initially in the Synthetic Feature Enhancement Module. Subsequently, the Multi-Scale Feature Interaction Module generates multi-scale features for different scales of detection and localization tasks. Finally, the Detection and Localization Module obtains predicted labels and region masks.

\subsection{Forgery Feature Enhancement Module}
Different from previous methods that utilize image noise residue obtained from SRM\cite{name34}, we first use a noise extractor to learn a noise-sensitive fingerprint from RGB images. In addition, the use of multiple types of characteristics to participate in training is considered to be one of the effective means to improve the training effect. Therefore, in addition to using image noise features, we also introduce spatial domain RGB features of the image. As mentioned earlier, because there are significant frequency inconsistencies among images generated by different methods, we add a frequency feature extraction branch to obtain image frequency domain features, hoping to obtain a stronger feature representation of forged areas through the combination of multiple features.

To enhance the forgery features we extracted, we use a dual attention fusion module that adds a threshold function to deeply fuse the RGB features and noise features of the spatial domain, including Channel Attention (CA) and Position Attention (PA), as shown in Figure 3. CA focuses on the relationships between all channels and adjusts channel weights accordingly, which helps achieve more accurate segmentation results. PA enhances the expressive power of features by establishing rich contextual relationships among local features based on the correlation between two-point features. Since features of different channels contribute differently to the detection task, we want to filter out weakly correlated features and enhance semantically correlated features. Therefore, we superimpose a threshold filter function on the channel weight of the output of the channel attention module to filter the channel.besides,We set a learnable weighted fusion method to fuse the output features of CA and PA, establishing connections between the context of the image and different positions of the features.Given input image features, they pass through both the Position Attention Module and Channel Attention Module. Adaptive fusion with the original image features results in new image features.

\subsection{Multi-Scale Feature Interaction Module}
Since the tampered regions of images generated by generative models vary in size, using a feature map with a fixed resolution may result in information loss. Therefore, we utilize feature maps of different resolutions to capture richer local and global features. Conventional network structures concatenate different resolutions, but we adopt a lightweight backbone network, HRNet\cite{name35}, for interactive fusion between different scale features, which is a parallel feature interaction method. Specifically, first, we extract features of the given input image through the Forgery Feature Enhancement Module. Then, we set up four branches to obtain feature maps of four resolutions, denoted as 
\(\theta_b\),where .Each branch can extract feature maps of specific resolutions, and thorough feature interaction is performed among different branches by establishing full connections. The output of each scale branch is fused from the outputs of all branches. For example, the downsampling 4x branch 
\begin{math}
	\theta_{3}
\end{math}  is obtained by downsampling branch \begin{math}
	\theta_{1}
\end{math}  by 4x, branch 
\begin{math}
	\theta_{2}
\end{math} by 2x, upsampling branch 
\begin{math}
	\theta_{4}
\end{math} by 2x, and adding the output of \begin{math}
	\theta_{3}
\end{math} itself.

\subsection{ Progressive Detection and Localization Module}
\begin{figure}[h]
	\centering
	\includegraphics[width=\linewidth]{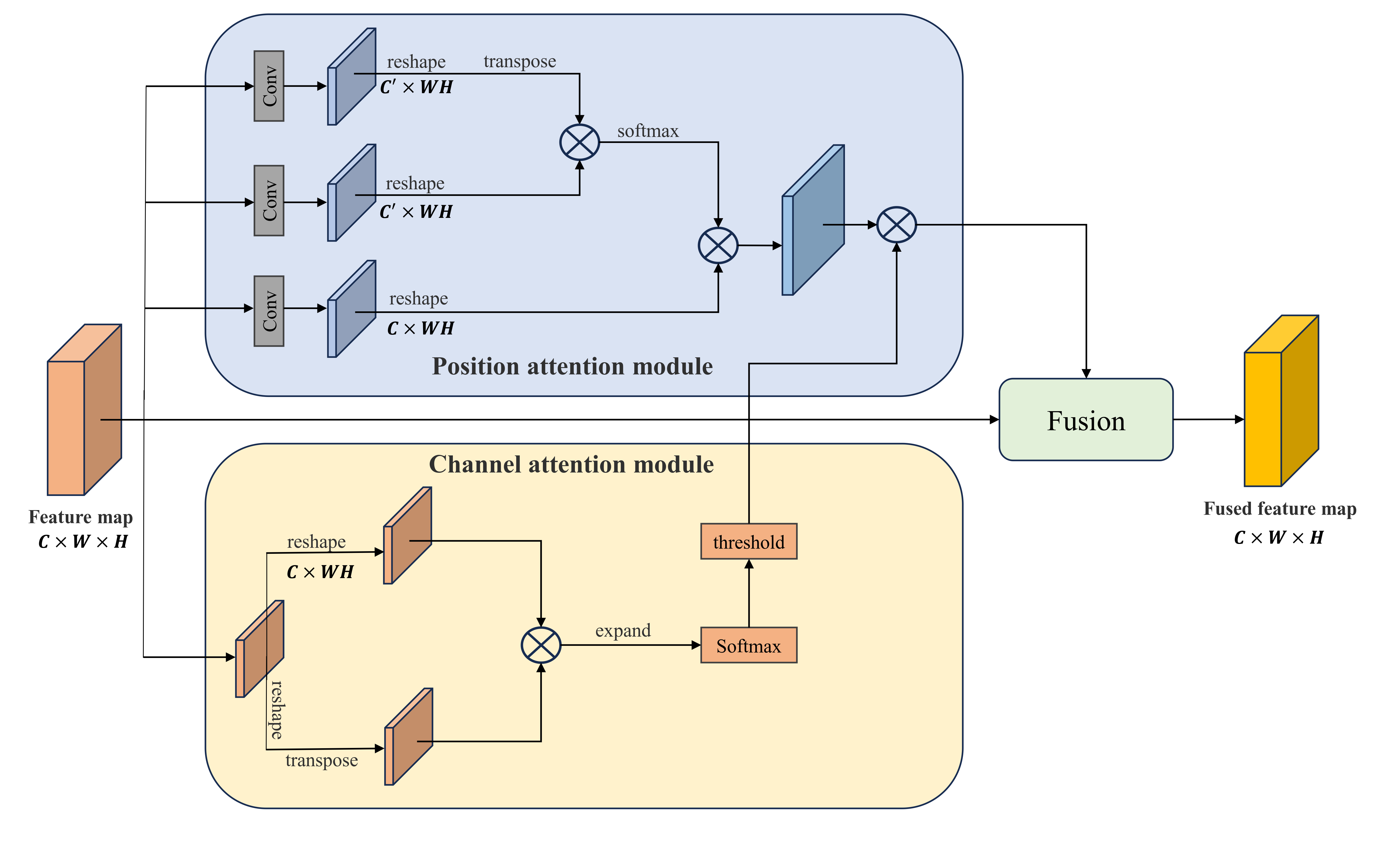}
	\caption{Dual Attention Feature Fusion Module. }
\end{figure}
Inspired by previous works\cite{name29,name36}, we use a hierarchical network to achieve coarse-to-fine image-level forgery detection and pixel-level forgery localization. Specifically, first, we obtain feature maps of different resolutions as described above. Then, starting from the lowest resolution feature map, we perform category prediction and region prediction for each level in a detection module and a localization module, respectively. In the detection module, we use a simple linear classifier, while in the localization module, we adopt a self-attention mechanism to map the feature maps of each level to a
\begin{math}
	mask_{b}
\end{math}.

We establish correlations between different hierarchical branches. Specifically, after obtaining the results of the current level, in the next level prediction process, we introduce the classification and localization prediction results of the previous level as prior knowledge to constrain the results of the current level. We use a progressive network to iteratively improve our detection and localization results. For the detection task, considering that the lower resolution features may suffer from information loss and fail to achieve fine classification, we set relatively rough classification categories for them. Similarly, for the localization task, we utilize the coarse estimate of forged areas obtained from the low-resolution branch to guide the improvement of localization in the next level. The above process can be described as follows: given an image 
\begin{math}
	X\in \mathbb{R}^{H\times W\times 3}
\end{math}   
, the features extracted from the four branches are denoted as 
\begin{math}
	F_{1}\in \mathbb{R} ^{H\times W\times C} 
\end{math}, 
\begin{math}
	F_{2}\in \mathbb{R} ^{H/s{} \times W/s\times sC} 
\end{math},
\begin{math}
	F_{3}\in \mathbb{R} ^{H/s^{2} {} \times W/s^{2} \times s^{2} C}      
\end{math},
and 
\begin{math}
	F_{4}\in \mathbb{R} ^{H/s^{3} {} \times W/s^{3} \times s^{3} C}         
\end{math}. Then, we obtain the forged area localization as:
\begin{equation}
	mask_{b}=f_{loc}^{b}(F_{b}(X) )     
\end{equation}
\begin{equation}
	mask_{b+1}=softmax(F_{b+1}+ \Gamma (mask_{b}\cdot F_{b+1}/2  ))          
\end{equation}
where 
\begin{math}
	f_{loc}^{b}(\cdot )       
\end{math} 
represents the localization prediction head of branch b, 
\begin{math}
	\Gamma (\cdot )      
\end{math}  represents a linear interpolation operation.
We denote the category prediction results and prediction probabilities of different branches as \begin{math}
	\theta_{b}        
\end{math} 
and 
\begin{math}
	p(y_{b}|X )        
\end{math} , respectively. Then, we have:
\begin{equation}
	p(y_{b}|X )=softmax(f_{det}^{b}(F_{b}(X) ) )    
\end{equation}
\begin{equation}
	p(y_{b+1}|X )=softmax(\theta _{b+1}(X)\odot (1+p(y_{b}|X ) ))   
\end{equation}

where 
\begin{math}
	f_{det}^{b}(\cdot )        
\end{math}  
represents the category detection head on branch b.

\subsection{ Loss Function}
To train our DA-HFNet method, we set different loss functions for the detection and localization tasks on each branch as optimization objectives. Specifically, for the detection task of forged categories, we define the optimization objective on each branch as:
\begin{equation}
	\mathfrak{L}_{det}^{b}(X)=-\frac{1}{N}\sum y_{i}\cdot log(p(y_{b}|X ))     
\end{equation}
Here, N represents the number of samples, \begin{math}
	y_{i}       
\end{math} 
represents the true class labels, and 
\begin{math}
	p(y_{b}|X )      
\end{math} 
epresents the predicted category probability on branch b. Therefore, the classification loss for all four branches is:
\begin{equation}
	\mathfrak{L} _{det}(X)=\sum_{b=1}^{4}\mathfrak{L}_{det}^{b}(X)      
\end{equation}

For the localization task of forged regions, we utilize the binary cross-entropy loss function to define the optimization objective on each branch as:

\begin{equation}
	\mathfrak{L}_{loc}^{b}(X)=-\frac{1}{H^{b}W^{b}  }\sum_{i=1}^{H^{b}}\sum_{j=1}^{W^{b}}(y_{i,j}^{b}\cdot log(p_{i,j}^{b}(X) ) )         
\end{equation}
Here, 
\begin{math}
	H^{b}      
\end{math} and 
\begin{math}
	W^{b}      
\end{math} represent the length and width of the image on branch b, 
\begin{math}
	y_{i,j}^{b}      
\end{math} represents the true mask label at position (i,j) on branch b, and 
\begin{math}
	p_{i,j}^b(X)     
\end{math}
represents the predicted probability at position (i,j)on branch b. Consequently, the localization loss for all four branches is:
\begin{equation}
	\mathfrak{L}_{loc}(X)=\sum_{b=1}^{4}\mathfrak{L}_{loc}^{b}(X)        
\end{equation}
Since the localization of forged regions falls under the category of image segmentation, and considering that segmentation tasks often lead to misjudgments of forged and real pixels at boundary positions, and AI-edited images often have unclear boundary traces in edited areas, we introduce a learnable edge loss \begin{math}
	\mathfrak{L}_{edge}       
\end{math}. As we are more concerned about the final localization results of forged regions, we only calculate the edge loss on the highest-resolution branch. Our edge loss \begin{math}
	\mathfrak{L}_{edge}       
\end{math} is defined as:
\begin{equation}
	\mathfrak{L}_{edge}=w\times (mean(S_{x}(M)-S_{x}(m))+mean(S_{y}(M)-S_{y}(m)))        
\end{equation}
Here, w is a hyperparameter controlling the edge loss, \begin{math}
	S_{x}       
\end{math} and \begin{math}
	S_{y}       
\end{math} represent the Sobel convolution kernels in the x and y directions,mean indication of performing an average calculation. M represents the true mask image, and m represents the predicted mask image.

In summary, combining equations (6), (8), and (9), we use a combination of three losses as the total loss:
\begin{equation}
	Loss=\mathfrak{L}_{det}+\mathfrak{L}_{loc}+\mathfrak{L}_{edge}          
\end{equation}

\section{Experiment}
\begin{figure*}[h]
	\centering
	\includegraphics[width=\linewidth]{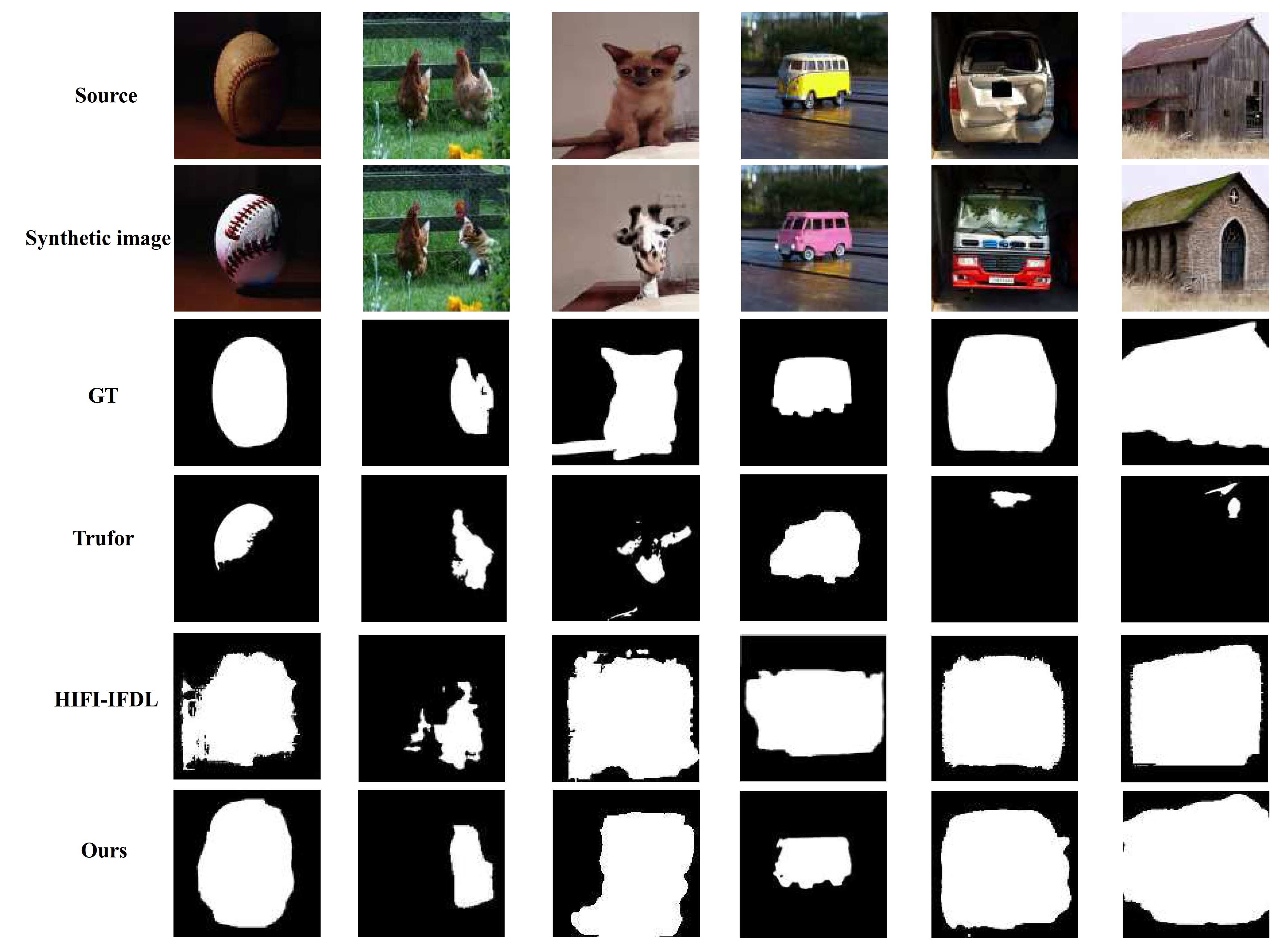}
	\caption{Comparison of Synthetic Region Localization Results.}
\end{figure*}
\subsection{Dataset and Experimental Setup}
Dataset: To address the current lack of corresponding datasets for some advanced image editing methods, we have constructed an image generation and editing dataset based on GAN and Diffusion Model. It includes both image and text modes in the guiding process. In Figure 1, we provide examples of some images from the dataset. We primarily constructed the DA-HFNet dataset for two reasons: 1) Generated images based on generative models are not only becoming more realistic in terms of quality but also diverse in terms of generation methods. Especially, some image editing methods no longer have clear editing boundaries. However, to our knowledge, existing datasets do not cover this aspect of research. 2) Detecting and locating forged images require studying various types of forgery methods, and a dataset containing various methods is necessary for research. 
\begin{table}[h]
	\caption{Composition Structure of the DA-HFNet Dataset}
	\resizebox{\linewidth}{!}{
		\begin{tabular}{cccccccccc}
			\hline
			\multirow{2}{*}{\textbf{Method}} & \multicolumn{2}{c}{\textbf{Generator Category}} &  & \multicolumn{2}{c}{\textbf{Forgery region}} &  & \multirow{2}{*}{\textbf{Guidance}} &  & \multirow{2}{*}{\textbf{Number}} \\ \cline{2-3} \cline{5-6}
			& GAN             & Diffusion            &  & Full           & Partial           &  &                           &  &                         \\ \hline
			BigGAN\cite{name37}                  & Y              & N                    &  & Y              & N                 &  & image                     &  & 3k                      \\
			DDPM\cite{name4}                    & N               & Y                    &  & Y              & N                 &  & image                     &  & 3k                      \\
			FuseDream\cite{name38}               & Y               & N                    &  & Y              & N                 &  & text                      &  & 3k                      \\
			GLIDE\cite{name20}                   & N               & Y                    &  & Y              & N                 &  & text                      &  & 3k                      \\
			Inpaint Anything\cite{name24}        & N               & Y                    &  & N              & Y                 &  & text                      &  & 3k                      \\
			Paint by Example\cite{name24}        & N               & Y                    &  & N              & Y                 &  & image                     &  & 3k                      \\
			StyleCLIP\cite{name25}               & Y               & N                    &  & N              & Y                 &  & text                      &  & 3k                      \\ \hline
		\end{tabular}
	}
\end{table}
\begin{table}[]
	\caption{Image-level forgery detection ACC score and F1 score.The best results are marked in bold, and the second-best results are underlined. 
		\begin{math}
			*       
		\end{math} means we apply authorreleased pre-trained models.}
	\resizebox{\linewidth}{!}{
		\begin{tabular}{ccccccccc}
			\hline
			\multirow{2}{*}{\textbf{Method}} & \multicolumn{2}{c}{\textbf{GAN-based}} & \textbf{} & \multicolumn{2}{c}{\textbf{DM-based}} & \textbf{} & \multicolumn{2}{c}{\textbf{AVG}} \\ \cline{2-3} \cline{5-6} \cline{8-9} 
			& \textbf{ACC(\%)}       & \textbf{F1(\%)}       & \textbf{} & \textbf{ACC(\%)}       & \textbf{F1(\%)}      & \textbf{} & \textbf{ACC(\%)}    & \textbf{F1(\%)}    \\ \hline
			\textbf{Trufor*}                 & 66.39              & 67.76             &           & 64.33              & 68.14            &           & 65.36           & 67.95          \\
			\textbf{PSCC-Net*}               & 50.86              & 61.03             &           & 47.62              & 55.85            &           & 49.24           & 58.44          \\
			\textbf{PSCC-Net}                & 95.56              & \textbf{99.13}             &           & 93.48              & \textbf{98.25}            &           & 94.52           & \textbf{98.69}          \\
			\textbf{HIFI-IFDL*}              & 66.46              & 49.15             &           & 64.16              & 45.89            &           & 65.31           & 47.52          \\
			\textbf{HIFI-IFDL}               & {\ul 96.37}        & 98.24             &           & {\ul 94.47}        & {\ul 98.05}      &           & {\ul 95.42}     & 98.13          \\
			\textbf{DA-HFNet(ours)}          & \textbf{99.07}              & {\ul 98.81}       &           & \textbf{99.63 }             & 97.93            &           & \textbf{99.35}          & {\ul 98.37}    \\ \hline
		\end{tabular}
	}
\end{table}
The composition structure of our dataset is shown in Table 1. Image-guided generated images come from BigGAN\cite{name37}, DDPM\cite{name4}, and Paint by Example\cite{name24}. BigGAN is a representative GAN-based whole image generation method; DDPM is a representative diffusion model-based whole image generation method; Paint by Example can use the diffusion model to draw a given image into another image in a specified area. Text-guided generated images come from FuseDream\cite{name38}, GLIDE\cite{name20}, Inpaint Anything\cite{name5}, and StyleCLIP\cite{name25}. FuseDream can generate images that match the description by relying on GAN from a given text segment; GLIDE can generate images that match the description by relying on the diffusion model from a given text segment; Inpaint Anything uses the diffusion model to change the specified area in a real image into an image that matches the text description; StyleCLIP edits images into images that match the text description by relying on GAN. For real images, 10k images are randomly selected from COCO2017 and ImageNet, respectively. 
\begin{table*}[!h]
	\caption{Verify experimental results across datasets.The best results are marked in bold, and the second-best results are underlined. }
	\resizebox{\linewidth}{!}{
		\begin{tabular}{clllllllllll}
			\hline
			\multirow{3}{*}{\textbf{Method}} & \multicolumn{5}{c}{\textbf{CoCoGLIDE dataset}}                                                                                                                                         &                      & \multicolumn{5}{c}{\textbf{HIFI-IFDL dataset}}                                                                                                                                                  \\ \cline{2-12} 
			& \multicolumn{2}{c}{\textbf{Detection}}                                     & \multicolumn{1}{c}{} & \multicolumn{2}{c}{\textbf{Localization}}                                  & \multicolumn{1}{c}{} & \multicolumn{2}{c}{\textbf{Detection}}                                     & \multicolumn{1}{c}{\textbf{}} & \multicolumn{2}{c}{\textbf{Localization}}                                  \\ \cline{2-3} \cline{5-6} \cline{8-9} \cline{11-12} 
			& \multicolumn{1}{c}{\textbf{ACC(\%)}} & \multicolumn{1}{c}{\textbf{F1(\%)}} & \multicolumn{1}{c}{} & \multicolumn{1}{c}{\textbf{ACC(\%)}} & \multicolumn{1}{c}{\textbf{F1(\%)}} & \multicolumn{1}{c}{} & \multicolumn{1}{c}{\textbf{ACC(\%)}} & \multicolumn{1}{c}{\textbf{F1(\%)}} & \multicolumn{1}{c}{}          & \multicolumn{1}{c}{\textbf{ACC(\%)}} & \multicolumn{1}{c}{\textbf{F1(\%)}} \\ \hline
			\textbf{PSCC-Net}                & 42.59                                & {\ul 41.73}                         &                      & 60.35                                & 42.82                               &                      & 79.26                                & 71.34                               &                               & 67.42                                & 73.31                               \\
			\textbf{HIFI-IFDL}               & {\ul 49.58}                          & 38.24                               &                      & {\ul 71.06}                          & {\ul 51.27}                         &                      & \textbf{86.18}                       & {\ul 82.13}                         &                               & {\ul 72.24}                          & {\ul 82.68}                         \\
			\textbf{DA-HFNet(ours)}          & \textbf{51.06}                       & \textbf{47.39}                      &                      & \textbf{74.48}                       & \textbf{58.46}                      &                      & {\ul 84.20}                          & \textbf{85.42}                      &                               & \textbf{73.09}                       & \textbf{83.56}                      \\ \hline
		\end{tabular}
	}
\end{table*}

Evaluation Metrics: Drawing from previous work experience, we selected Accuracy Score (ACC) and F1 Score (F1) as the evaluation metrics for our experiments. 

Experimental Settings: DA-HFNet is implemented in PyTorch and trained on four NVIDIA 3090 GPUs. The size of the input images is set to 256×256. The initial learning rate is 0.0002 and periodically decays to 1e-08. The training is conducted for 50 epochs. We split the training data and the test data in a 9:1 ratio.We apply conventional data augmentation methods during the training process, including rotation and flipping.
\subsection{Forgery  Image Detection and Localization}
We tested our model on the DA-HFNet dataset constructed in this paper. First, we compared our method with baseline methods in image-level classification and pixel-level localization. Table 2 reports the results of different methods in image-level attribute category classification. Specifically, we observed that the performance of detectors pre-trained on our dataset was generally lower, which is directly related to the feature representation capability of the learned detector parameters. Trufor utilizes features containing noise and RGB features; PSCC-Net uses RGB features of the image and employs a hierarchical structure. HIFI-IFDL uses RGB features and frequency domain features, also utilizing a hierarchical structure. Both PSCC-Net and HIFI-IFDL are representative methods based on hierarchical networks. Detectors trained on our dataset perform the best on counterfeit images based on GAN and diffusion model, showing the highest ACC scores and competitively high F1 scores.
Next, we report the performance of different detectors on pixel-level counterfeit region localization in Table 4. Our baseline model adopts the detector used in image-level counterfeit category classification. From Table 4, we observe that pre-trained detectors do not perform well on our dataset. This is partly because pre-trained PSCC-Net and HIFI-IFDL primarily segment objects from artificially edited counterfeit images, which have significant fingerprint differences from AIGC-edited counterfeit images. We then trained PSCC and IFDL detectors on our DA-HFNet dataset, resulting in noticeable improvements in both ACC and F1 scores. Additionally, we observed that for counterfeit region localization, our method achieved the best performance in both ACC and F1 scores, with an average ACC score increase of 3.93\% and an average F1 score increase of 2.72\% compared to the second-best method. We showcase our counterfeit region localization results and comparisons with other methods in Figure 4.It can be observed that our method achieves better positioning results.

\begin{table}[]
	\caption{pixel-level forgery localization ACC score and F1 score.The best results are marked in bold, and the second-best results are underlined. 
		\begin{math}
			*       
		\end{math} means we apply authorreleased pre-trained models.}
	\resizebox{\linewidth}{!}{
		\begin{tabular}{ccccccccc}
			\hline
			\multirow{2}{*}{\textbf{Method}} & \multicolumn{2}{c}{\textbf{GAN-based}} & \textbf{} & \multicolumn{2}{c}{\textbf{DM-based}} & \textbf{} & \multicolumn{2}{c}{\textbf{AVG}} \\ \cline{2-3} \cline{5-6} \cline{8-9} 
			& \textbf{ACC(\%)}       & \textbf{F1(\%)}       & \textbf{} & \textbf{ACC(\%)}      & \textbf{F1(\%)}       & \textbf{} & \textbf{ACC(\%)}    & \textbf{F1(\%)}    \\ \hline
			\textbf{Trufor*}                 & 78.15              & 79.32             &           & 76.49             & 77.50             &           & 77.32           & 78.41          \\
			\textbf{PSCC-Net*}               & 59.24              & 51.34             &           & 54.42             & 55.18             &           & 56.83           & 53.26          \\
			\textbf{PSCC-Net}                & 88.32              & 59.36             &           & 86.96             & 56.72             &           & 87.64           & 58.04          \\
			\textbf{HIFI-IFDL*}              & 66.41              & 41.94             &           & 60.13             & 42.82             &           & 63.27           & 42.38          \\
			\textbf{HIFI-IFDL}               & {\ul 91.25}        & {\ul 90.11}       &           & {\ul 88.49}       & {\ul 88.37}       &           & {\ul 89.87}     & {\ul 89.24}    \\
			\textbf{DA-HFNet(ours)}          & \textbf{93.47}     & \textbf{91.46}    &           & \textbf{91.25}    & \textbf{92.46}    &           & \textbf{92.36}  & \textbf{91.96} \\ \hline
		\end{tabular}
	}
\end{table}

	To verify the validity of our method, we performed validation experiments across data sets. The training dataset consists of our DA-HFNet dataset, and the verification dataset selects CoCoGLIDE and HIFI-IFDL datassets. CoCoGLIDE is a small image editing dataset  made by GLIDE model, which contains 512 forged images in total. HIFI-IFDL dataset is a comprehensive dataset containing multiple types of image forgery methods, from which we only select 500 images for each method synthesized by CNN. We report the results of our experiments for cross-dataset validation in Table 3.As can be observed from Table 3, our approach still shows obvious advantages over the baseline approach in cross-dataset validation tasks. However, it is undeniable that the accurate detection and location of unknown forgery methods is still a problem worthy of further research.


\subsection{Ablation Experiments}
We further conducted ablation experiments by comparing the ablation of the multi-feature dual attention interaction module in the method. Table 5 presents the results of the ablation experiments.
First, we eliminated the dual attention feature fusion module from the method and only used feature concatenation to fuse features. The method showed a decrease of 2.84\% and 6.2\% in ACC and F1, respectively, for counterfeit class detection, and a decrease of 3.08\% in ACC for counterfeit region localization, with a slight decrease in F1 score. We believe this is due to insufficient feature fusion after eliminating the dual attention fusion module.
Subsequently, we individually ablated the features used in the method. Rows 2-4 in Table 5 show that eliminating the image features from one branch resulted in varying degrees of performance degradation in the method. The most severe was the elimination of the frequency branch, where the method experienced a 7.68\% decrease in ACC for counterfeit class classification and a 5.75\% decrease in ACC for counterfeit region localization. Therefore, we believe that each branch's features positively contribute to the method's performance.
Finally, we replaced our noise features with noise residuals obtained using SRM filtering. The results, as shown in row 5 of Table 5, indicate that the detector trained using SRM filtering exhibited a decrease in both ACC and F1 metrics compared to our method.
\begin{table}[h]
	\caption{Ablation experiment results.RGB, Noise, and Frequency represent feature extractors. The optimal results are indicated in bold.}
	\resizebox{\linewidth}{!}{
		\begin{tabular}{cccccc}
			\hline
			\multirow{2}{*}{\textbf{Components of the Method}} & \multicolumn{2}{c}{\textbf{Detection}} & \textbf{} & \multicolumn{2}{c}{\textbf{Localization}} \\ \cline{2-3} \cline{5-6} 
			& \textbf{ACC(\%)}   & \textbf{F1(\%)}   & \textbf{} & \textbf{ACC(\%)}     & \textbf{F1(\%)}    \\ \hline
			\textbf{RGB, Noise, Frequency}                     & 96.51              & 92.17             &           & 89.28                & 91.37              \\
			\textbf{Noise, Frequency, DAM}                     & 95.55              & 97.34             &           & 88.43                & 93.74              \\
			\textbf{RGB, Noise, DAM}                           & 91.76              & 95.21             &           & 86.61                & 92.01              \\
			\textbf{RGB, Frequency, DAM}                       & 92.84              & 97.81             &           & 86.63                & 91.17              \\
			\textbf{RGB,SRM, Frequency, DAM}                   & 96.01              & 96.12             &           & 88.55                & 90.90              \\
			\textbf{RGB, Noise, Frequency, DAM}                & \textbf{99.35}     & \textbf{98.37}    &           & \textbf{92.36}       & \textbf{91.96}     \\ \hline
		\end{tabular}
	}
\end{table}

In subsequent experiments, we studied the impact of edge loss on the experiment and reported the corresponding results in Table 6. It can be observed from Table 6 that when we remove the edge loss, the ACC score and F1 score of the model in the forgery category classification task decrease by 6.47\% and 4.88\% respectively, and the ACC score and F1 score in the forgery area localization task decrease by 4.35\% and 5.79\% respectively. This proves the positive effect of the edge loss we set on the experiment.

\begin{table}[h]
	\caption{Comparison of performance with and without edge loss.}
	\resizebox{\linewidth}{!}{
		\begin{tabular}{cccccc}
			\hline
			\multirow{2}{*}{\textbf{Components of the Method}} & \multicolumn{2}{c}{\textbf{Detection}} & \textbf{} & \multicolumn{2}{c}{\textbf{Localization}} \\ \cline{2-3} \cline{5-6} 
			& \textbf{ACC(\%)}   & \textbf{F1(\%)}   & \textbf{} & \textbf{ACC(\%)}     & \textbf{F1(\%)}    \\ \cline{1-3} \cline{4-6} 
			\textbf{DA-HFNet(No edgeloss)}                     & 92.88              & 93.49             &           & 88.07                & 86.17              \\
			\textbf{DA-HFNet}                                  & \textbf{99.35}     & \textbf{98.37}    & \textbf{} & \textbf{92.36}       & \textbf{91.96}     \\ \hline
		\end{tabular}
	}
\end{table}

\section{Conclusion }
 We proposed a progressive network based on multi-feature fusion for detecting and localizing high-quality forged images. Our method utilizes noise features more sensitive to noise and combines RGB and frequency features of images to capture richer forgery traces. Through the dual-attention fusion mechanism, we adaptively fuse multi-modal image features, and using a multi-branch feature interaction network, we achieve information exchange between features of different resolutions. Additionally, we constructed a dataset containing high-quality forged images generated by GAN and Diffusion models using text or image guidance. Extensive experiments on this dataset demonstrated the effectiveness of our method for detecting and localizing high-quality forged images.

\end{document}